# Hybrid Adversarial Spectral Loss Conditional Generative Adversarial Networks for Signal Data Augmentation in Ultra-precision Machining Surface Roughness Prediction

**S. Y. Shang [1], C. F. Cheung [1, *] and P. Zheng [1]**

[1] The Hong Kong Polytechnic University, State Key Laboratory of Ultra-Precision Machining Technology, Hong Kong, China

Tel.: +86 90455327, e-mail: benny.cheung@connect.polyu.hk

**Summary:** Intelligent surface roughness prediction in ultra-precision machining is crucial for real-time surface quality control. However, such datasets are often small and time-consuming to collect and annotate, hindering the prediction ability of prediction models. This paper presents a novel Hybrid Adversarial Spectral Loss Convolutional Conditional Generative Adversarial Network(HAS-CGAN) framework for data augmentation in ultra-precision machining (UPM) surface roughness prediction, addressing the critical challenge of limited training data. We systematically compare five CGAN variants, demonstrating that our proposed HAS-GAN outperform other CGANs for 1D force signal generation no matther on visual comparasion or quantative computation results, especially for signal with higher frequency. Data augmentation results show our proposed HAS-CGAN model significantly improves signal fidelity in higher frequency through Fourier-domain error penalization, achieving wavelet coherence lager than 0.85. Then, generated signals are incorporated with machining parameters as new model input. To compare the prediction accuracy before and after signal data augmentation, traditional machine learning methods (SVR, RF and LSTM) take hand-made features as input and several end-to-end models, such as BPNN, 1DCNN and CNN-Transformer take signals as input are utilized for ultra-precision milling surface roughness prediction.. Surface roughness prediction results show that adding more generated samples in the training dataset can improve the prediction accuracy for models that have automatic feature extraction ability compared to taking the original 52 real samples as model input. In detail, adding 520 or more generated samples can reach the best prediction ability for this dataset with a mean absolute percentage error of around 9%, while the original 52 real samples only have a prediction accuracy of 31.4%, addressing the critical challenge of limited training data.

**Keywords:** Signal data augmentation, Ultra-precision Machining Surface Roughness Prediction, Conditional Generative Adversarial Network, Customized loss.

## 1. Introduction

The attainment of excellent surface quality is an indicator of substantial cutting performance and currently serves as a characteristic hallmark of UPM technology [1][2]. Meanwhile, Ultra-precision surfaces are crucial for the functionality of numerous products and systems; in many cases, they are decisive in determining product performance [3]. For example, the quality of the surfaces within the optical tower of a lithography wafer stepper, which significantly dictates the small feature size of transistors within an integrated circuit [4]. Additionally, the ultra-precision smooth surfaces such as those needed for state-of-the-art optical devices, such as lenses and mirrors, are designed to perform specific functions like optimizing the transmission and reflectance of light [4]. Another form of ultra-precision surface is micro-grooved films that have various of applications such as the orientation of liquid crystals in LCD, plasma and OLED flat panel displays [5].

Physical measurement of the surface quality is typically the offline measurement using precision measuring instruments, such as ZYGO laser interferometers with high-precision surface topography metrology systems. To circumvent issues related to repositioning errors and time-consuming workpiece rotation, on-machine surface measurement approaches defined as measuring the surfaces without the removal of the workpiece from the machine tool, have been developed to enable timely precise detection [6][7]. However, both offline measurement and emerging online measurement systems share the fundamental limitation of only evaluating workpieces after machining completion, thereby preventing real-time understanding of the machining process and precluding immediate quality control interventions. In 2004，Cheng [8] proposed the innovative application of virtual metrology (VM) as a strategic approach to enhance overall equipment effectiveness in the semiconductor industry. VM is formally defined as: A method to conjecture operation performance of a process tool based on data sensed from the process tool and without physical metrology operation". Following this conceptual breakthrough, Cheng and his research team successfully applied VM methodology across diverse industry domains. Notably, applications include thickness prediction of chemical vapor deposition process in semiconductor manufacturing [9], automated dynamic-balancing inspection scheme for wheel machining in mechanical engineering [10], and the total quality inspection of sizing percentage, tensile strength and tensile modulus of spins in carbon fiber manufacturing [11]. VM establishes the theoretical and practical feasibility of the real-time prediction of UPM surface roughness, thereby enabling the realization of online monitoring and quality control for UPM processes. In practice, numerous researchers have investigated surface roughness prediction in UPM based on data acquired from ultra-precision machining processes [12][13][14]15].

A well-trained machine learning-based surface roughness prediction model offers dual advantages: significant reduction in manpower and time costs associated with physical measurements, and the capability to perceive and subsequently enable real-time control of surface quality [16]. The dataset for the prediction model is usually composed of signal data appearing in times series collected during the machining process in conjunction with machining parameters as model input and real-measured surface roughness as model output [6][7][17]. However, the inherent low efficiency of ultra-precision machining processes presents substantial challenges - both the collection of sufficient sensor data and the annotation of measurement data require considerable time and effort to train high-accuracy models. Consequently, such datasets in ultra-precision machining are often small in size [14][18][19], which significantly constrains the predictive capability of developed models. The primary objective of data augmentation in this context is to systematically increase the volume, quality, and diversity of available training data, in order to acquire a well-trained high high-accuracy model under a limited dataset.

Conventional data augmentation techniques, including Add Noise, Time Stretch, Pitch Shift, Mixup, and SpecAugment, can effectively expand raw signal datasets to improve prediction or classification accuracy [20][21][22][23]. More advanced, generative models appear particularly promising for addressing the low prediction accuracy problem caused by a low-volume training dataset in various real-world scenarios. For example, variant autoencoders and their variants are widely used for data augmentation. Wei-Ning Hsu et al. achieved about 35% absolute world error rate reduction on two speech recognition sets by the proposed augmentation method, which studies the latent representations obtained from VAEs that enable to transformation of nuisance attributes of speech through modifying the latent variables [24]. Or Deep Boltzmann Machines (DBMs), Salakhutdi a novel approach using DBMs to generate synthetic speech features that preserve the statistical properties of real data for improving the robustness of speech recognition systems [25].

Generative Adversarial Nets (GAN) were initially introduced by Goodfellow et.al. as a groundbreaking framework for estimating generative models via an adversarial process to generate artificial images. And it has been proven that their generated samples are better than samples generated by existing methods [26]. Also, Liu et al use GAN to directly predict synchronized raw audio signals and generate realistic sounds from video in real time, which provides a viable solution for applications such as sound design and dubbing [27]. Compared with variational autoencoders, GANs do not introduce any deterministic bias. Variational methods introduce deterministic bias because they optimize the lower bound of the log-likelihood rather than the likelihood itself, which appears to result in GANs producing sharper generated instances than VAEs. In contrast to VAEs, GANs do not have a variational lower bound [26]; if the discriminator is well-trained, the generator can perfectly learn the training sample distribution. In other words, GANs are asymptotically consistent [28], whereas VAEs are biased. However, methods relying on Markov chain Monte Carlo sampling suffer from low computational efficiency [29], sensitivity to hyperparameters that make tuning difficult, and challenging optimization problems. Furthermore, similar to VAEs, the generated samples typically exhibit greater blurring and higher noise levels compared to GANs.

Comprehensive empirical studies have consistently shown that the images produced by GANs exhibit higher fidelity and realism, making them suitable for applications in computer vision and beyond [30][31]. On the CIFAR-10 dataset, the Inception Score of GANs was 22% higher than that of VAEs and quantitative evaluation demonstrated that VAEs performed significantly worse than GANs in terms of Fréchet Inception Distance [32]. However, the original GAN is not suitable for surface roughness prediction since it can only generate signals without surface roughness labels. Conditional generative adversarial networks (CGANs) leverage an additional surface roughness label for both the discriminator and generator. Therefore, sensor signals generated by CGANs are accompanied by labels, that can be directly used for training a more robust prediction model. However, CGANs also suffer from training instability, gradient vanishing, and mode collapse issues [33][34]. In Auxiliary Classifier Generative Adversarial Networks (ACGANs), the classifier loss function mitigates mode collapse by providing additional gradient signals [35]. Similarly, Wasserstein Conditional Generative Adversarial Networks (WCGANs) fundamentally resolve the gradient vanishing issue by replacing JS divergence with Wasserstein distance (also known as Earth-Mover distance) [36]. Also, it can prevent the generator from converging to suboptimal local minima by adding the Lipschitz constraint (enforced via gradient penalty or weight clipping) [37]. Given these complementary strengths, both ACGANs and WCGANs are included in our current study as comparative studies.

The application of CGANs and their derivatives spans various domains, including generating minor class data in induction motor fault diagnosis [38], class-specific synthetic time-series sequences of arbitrary length [39], and highly realistic emoji images [40]. Despite this broad applicability, remarkably little research attention has focused on the potential applications of CGANs in ultra-precision machining scenarios where both sensor data and labelled data are exceptionally scarce that adversely affects the model's prediction accuracy. Moreover, data collection in ultra-precision machining environments presents substantially greater challenges compared to internet-based applications with large user bases. Consequently, we novelly propose HAS-CGAN that adds spectral loss into original CGAN loss to improves signal fidelity in higher frequency through Fourier-domain error penalization. Our experimental results conclusively demonstrate that appropriate use of HAS-CGAN can most effectively address the low accuracy problem caused by severely limited ultra-precision machining datasets.

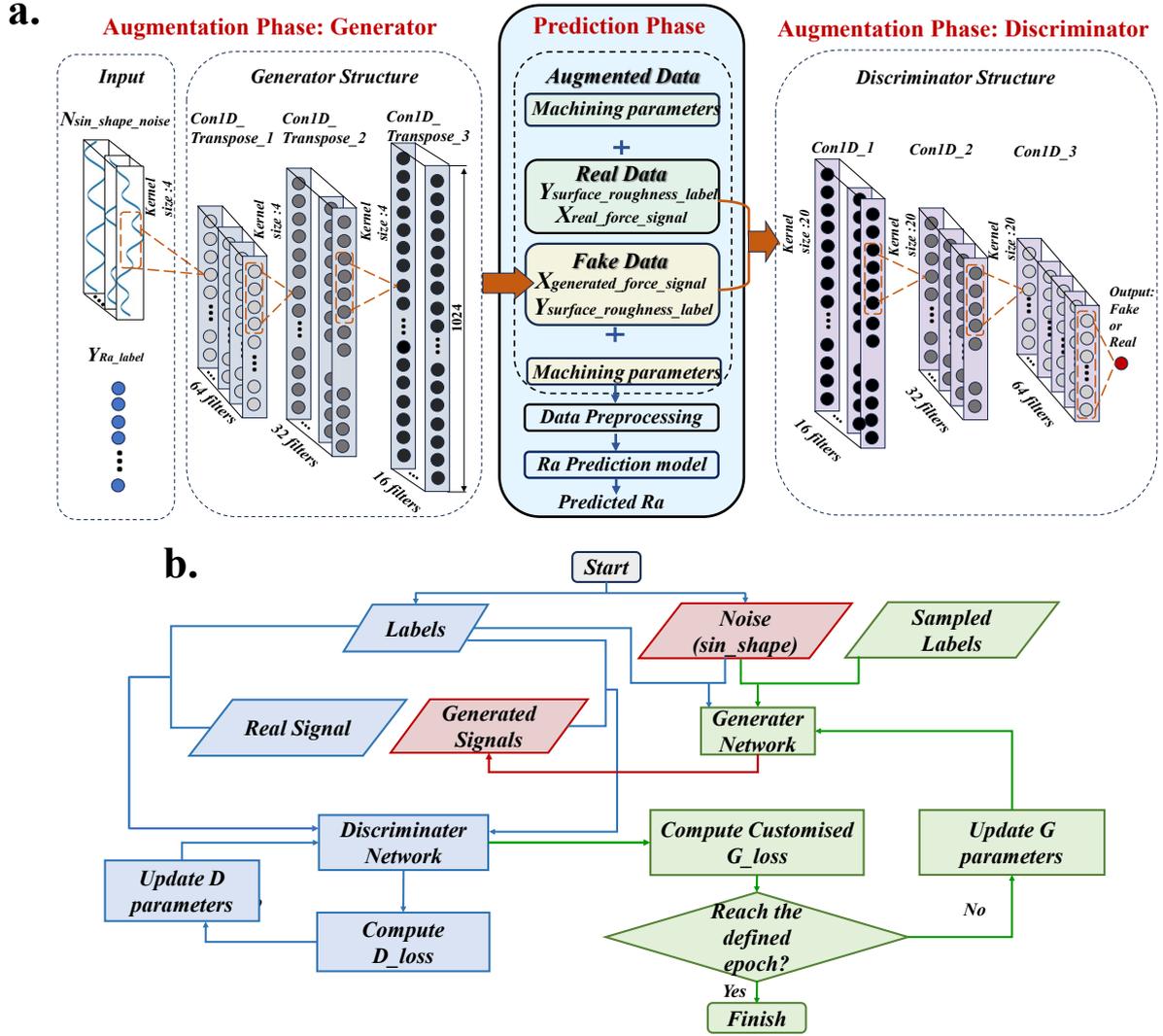

**Fig. 1.** Schematic illustration of Conditional Generative Adversarial Networks for signal data augmentation in ultra-precision surface roughness prediction. **a.** The overall architecture of our proposed HAS-CGAN for signal data augmentation in ultra-precision surface roughness prediction. **b.** The flowchart of the HAS-GAN training process for each epoch.

## 2. System framework and methodology
### 2.1 System framework

The overall architecture of the proposed 1D-Convolutional Conditional Generative Adversarial Network with customized loss is illustrated in **Fig. 1. a**, comprising two phases: an augmentation phase, which contains the generator and the discriminator, and a subsequent prediction phase.

The specific structures and settings of hyperparameters for the generator and discriminator are adapted from reference[26]. In the generator architecture, three 1D-convolutional transpose layers are implemented by taking sinusoidal noise with 100 latent dimensions combined with surface roughness labels as input. The selection of sinusoidal noise is theoretically justified by the periodic nature of force signals, which, according to Fourier analysis, can be decomposed into multiple sinusoidal components with varying amplitudes and frequencies. This design choice enables the generator to more effectively mimic realistic signal patterns. The three convolutional layers employ filter sizes of 64, 32 and 16 respectively with a uniform kernel size of 20 selected to match the approximate periodicity observed in real signals, thereby enhancing both feature learning capability and method interpretability. The final generator output consists of synthetic force signals with corresponding surface roughness labels.

Conversely, the discriminator implements a three-layer 1D convolutional neural network for binary classification of input signals as real or generated. The filter sizes follow an inverted sequence (16, 32, 64) relative to the generator, while maintaining the same 20-unit kernel size for consistent periodic feature extraction. The discriminator processes both real and generated signals along with their corresponding labels, outputting a probability estimate of input authenticity.

Following complete CGAN training, the well-optimized generator can produce numerous synthetic signals that closely resemble real experimental data.

The prediction phase serves to evaluate whether generated signals can meaningfully enhance surface roughness prediction accuracy. This phase incorporates three key components: augmented datasets (combining limited real signals with abundant generated samples), data preprocessing modules, and surface roughness prediction models. Augmented data has both little real signals with corresponding machining parameters and labels and more generated signals with corresponding machining parameters and labels. The number of samples in the dataset after augmentation can be several times the original one. We evaluate two distinct prediction approaches: traditional signal processing methods involving manual extraction of time-domain and frequency-domain features followed by SVM, transformer, or neural network models; and deep learning methods (1D-CNN, DNN, transformers) with automated feature extraction capabilities. For traditional methods, data preprocessing follows established procedures from reference [41], while deep learning approaches employ simpler preprocessing limited to anomaly removal and data segmentation.

**Fig.1.b** details the training methodology of HAS-CGAN. Unlike conventional predictive neural networks, CGANs require a separate two-stage training of discriminative and generative components to maintain adversarial equilibrium. In **Fig.1.b**, the discriminator (processes painted blue) is first trained and the generator (processes painted green) is trained secondly. Red parts both occur in the training process of the discriminator and the generator.

In the first phase, noise with sinusoidal noise and random generated labels are input of the generator network and output generated signals directly without hyperparameters updating. Then, generated signals and real signals with their corresponding labels are the input of the discriminator network, followed by the computed discriminator loss whose goal is to distinguish the generated signals and real signals. If this epoch is not the last one defined, the hyperparameters of the discriminator will be updated through the Adam optimizer. Till now, training for the discriminator in one epoch is finished.

In the second phase, the HAS-CGAN network conducts training of the generator. With input of noise and randomly sampled labels, the generator gives out the generated signals and uses the above-trained discriminator with its parameters frozen in the second training process to compute customized generator loss, whose goal is to make generated signals more similar to the real ones. A customized loss (Hybrid Adversarial Spectral loss) is composed of traditional generator loss and an extra spectral loss to improve generation quality for the signals with relatively high frequency.

After several epochs of iterative training of the above two processes, the discriminator and generator reach a Nash Equilibrium**.** The discriminator becomes highly capable of signal authentication, while simultaneously the generator produces synthetic signals indistinguishable from real data to the discriminator.

### 2.2 Theory foundations of CGANs.
### 2.2.1 Hybrid Adversarial-Spectral CGAN (HAS-CGAN)

HAS-CGAN trains generator G and discriminator D oppositely to each other by leveraging additional surface roughness labels for both discriminator and generator [42]. In this paper, our proposed method is based on 1D-Convolutional CGAN, composed of a generator ($G$) with three 1D-Convolutional Transpose layers that is usually used in generative models and a discriminator ($D$) with three 1D-Convolutional layers that are used to distinguish real and fake signals. In the generator, the prior input noise $N(\mathbf{z})$, and surface roughness label $\mathbf{y}$ are combined in a joint hidden representation. Since mechanical signal data is periodic, thus, generated $sin$ function is taken as prior input noise $N(\mathbf{z})$. The feed forward of the first layer in the generator is shown in Eq. 1.

$$x(m)_1^i = \sum_{k=0}^{K-1} \omega(k) \cdot N(z)\left[\left\lfloor \frac{m-k\cdot s+p}{s} \right\rfloor\right] \quad (1)$$

Where, $x(m)_1^i$ is the output of the $m$-$th$ position of the $i$-$th$ signal for the first 1D-Convolutional Transpose layer, $\omega$ is the kernel, $K$ is the kernel size, $k$ is the $k$-$th$ element in the kernel, $s$ is the stride, $p$ is padding. Similarly, the output of the second 1D-Convolutional Transpose layer and the third one are illustrated in Eq.2 and Eq. 3.

$$x(m)_2^i = \sum_{k=0}^{K-1} \omega(k) \cdot x(m)_1^i \left[\left\lfloor \frac{m-k\cdot s+p}{s} \right\rfloor\right] \quad (2)$$

$$x(m)_{generated\_force\_signal}^i = \sum_{k=0}^{K-1} \omega(k) \cdot x(m)_2^i \left[\left\lfloor \frac{m-k\cdot s+p}{s} \right\rfloor\right] \quad (3)$$

Where, $x(m)_2^i$ is the output of the $m$-$th$ position of the $i$-$th$ signal for the second 1D-Convolutional Transpose layer and $x(m)_{generated\_force\_signal}^i$ the output of the $m$-$th$ position of the $i$-$th$ signal for the final generated signals.

Once got the output of the generator, the traditional loss function of the generator can be computed as Eq. 4.

$$Loss_1 = -\frac{1}{M}\sum_{i=1}^{M} D(label = real | (x_{generated\_force\_signal}^i, y^i)) \quad (4)$$

where, $M$ is the batch size, $x_{generated\_force\_signal}^i$ is the $i$-$th$ generated force signal data, $y^i$ is the corresponding surface roughness label. However, a customized loss is designed in Eq.5 by adding the spectral loss into the traditional loss function to enhance spectral-level fidelity in signal generation via frequency-domain constrained learning.

$$Loss_2 = \frac{1}{M}\cdot\frac{1}{T}\sum_{t=1}^{T}\left\||STFT(x_{real\_force\_signal}^i)| - |STFT(x_{generated\_force\_signal}^i)|\right\|_F^2 \quad (5)$$

Where, $M$ is the batch size, $T$ is the time frames, $STFT$ is the abbreviation of Short-Time Fourier Transform, $x_{real\_force\_signal}^i$ is the $i$-$th$ real force signal data, $|\cdot|$ is the magnitude of $STFT$. $\|\cdot\|_F$ is the square root of the sum of squared matrix elements.

Therefore, the final loss function of the generator for our proposed method is shown in Eq. 6 as follows.

$HAS\_Loss_G = \gamma_1 \cdot Loss_1 + \gamma_2 \cdot Loss_2$, constrained by $\gamma_1 + \gamma_2 = 1$ (6)

Where, $HAS\_Loss_G$ is the hybrid adversarial spectral loss we proposed, $\gamma_1$ and $\gamma_2$ are weight coefficients and the sum of them is 1. We propose HAS-CGAN, where the generator is optimized via a hybrid loss combining. adversarial training and spectral constraints. As illustrated in Section 2, with the discriminator's parameters frozen, the combined generator-discriminator structure is trained. The aim of $G$ is to minimize the generator loss $Loss_G$, forcing $G$ to produce samples that $D$ classifies as real, thereby improving generation quality.

In discriminator force signal signals $x$ and surface roughness labels $y$ are presented as input. As the inverse process of the generator, the discriminator is composed of three 1D-Convolutional layer. The number of filters increases layer by layer from 16 filters to 32 filters and to 64 filters. The filter size keep the same as 20. The output for the first layer, sceond layer and final layer in the discriminator are shown in Eq. 7, Eq. 8 and Eq. 9 resepectively.

$O(m)_1^i = \sum_{k=0}^{K} \omega[k] \cdot x^i[m \cdot s + k - p]$ (7)
$O(m)_2^i = \sum_{k=0}^{K} \omega[k] \cdot O_1^i[m \cdot s + k - p]$ (8)
$O(m)_3^i = \sum_{k=0}^{K} \omega[k] \cdot O_2^i[m \cdot s + k - p]$ (9)

Where, $O(m)_1^i$, $O(m)_2^i$, and $O(m)_3^i$ are outputs of $m$-$th$ position of the $i$-$th$ signal for the first layer, second layer and last layer of the discriminator. $\omega$ is the kernel, $K$ is the kernel size, $k$ is the $k$-$th$ element in the kernel, $s$ is the stride, $p$ is the padding. Therefore, the loss function of the discriminator is shown in Eq. 10.

$Loss_D = -(\frac{1}{M}\sum_{i=1}^{M} \log(D(label = real|(x_{real\_force\_signal}^i, y^i)) - (\frac{1}{M}\sum_{i=1}^{M} \log(1 - D(label = fake|(x_{generated\_force\_signal}^i, y^i))$ (10)

In each epoch, the discriminator is trained first on the current mini-batch, using both real signal samples $x^i \in x_{real\_force\_signal}^i$ and generated signal samples $x^i \in x_{generated\_force\_signal}^i$. The training goal of $D$ is to minimize $Loss_D$, which combines real sample loss and fake sample loss. This phase enhances $D$'s ability to distinguish between real and synthetic signals.

### 2.2.2 Theory of ACGAN and WCGAN

Similar to CGAN, an auxiliary classifier is incorporated into the standard GAN discriminator, jointly optimizing the authenticity of generated samples and their class labels to enhance both generation quality and categorical controllability [43] The loss function for ACGAN is slightly different that adding the classification loss shown in Eq. 11 and Eq. 12.

$L_G^{ac} = -E(logD_{real}(x_{generated\_force\_signal}) + \alpha E[logD_{class}(y|x_{generated\_force\_signal})]$ (11)

$L_D^{ac} = E[\log D_{real}(x_{real\_force\_signal})] + E[\log(1 - D_{real}(x_{generated\_force\_signal}))] - E[logD_{class}(y|x_{real\_force\_signal})]$ (12)

Where, $L_G^{ac}$ is the $G$ loss of ACGAN, $L_D^{ac}$ is the $D$ loss of ACGAN, $E$ is the expected value, $\alpha$ is the weight coefficient of classification loss, $D_{real}(\cdot)$ is the discriminator's authenticity output, $D_{class}(\cdot)$ is the auxiliary classifier's class probability output.

WCGAN is the integration of CGAN and Wasserstein GAN. Through Earth-Mover distance, Gradient penalty and condition adding, it can solve unstable training problems [44], especially popular for complex image generation problems. Its loss function uniquely computes the wasserstein distance and gradient penalty as shown in Eq. 13 and Eq. 14 and Eq. 15.

$L_G^{wc} = -E[D(x_{generated\_force\_signal})]$ (13)
$L_C^{wc} = E[D(x_{real\_force\_signal}|y)] - E[D(x_{generated\_force\_signal})] + \beta E[(\|\nabla_{\hat{x}} D(\hat{x}|y)\|_2 - 1)^2]$ (14)
$\hat{x} = \theta x_{real\_force\_signal} + (1 - \theta)x_{generated\_force\_signal}, \theta \epsilon U[0,1]$ (15)

Where, $L_G^{wc}$ is the $G$ loss of WCGAN, $L_C^{wc}$ is the critic loss of WCGAN, $\beta$ is the weight coefficient, $\nabla$ is the gradient. $\|\cdot\|_2$ is the L2 norm.

## 3. Experiment verification
### 3.1 Introducing the Dataset

To rigorously validate the effectiveness of our proposed HAS-CGAN (Hierarchical Attention-Supervised Conditional Generative Adversarial Network) data augmentation framework for ultra-precision machining surface roughness prediction, we conduct comprehensive experiments using the specialized dataset originally described in reference [14].

The experimental dataset comprises only 64 samples collected from ultra-precision milling operations, where each sample represents a complete set of machining process data containing: 1. Force signal data: High-frequency dynamic cutting force captured during the machining process. 2. Machining parameters (spindle speed, feed rate, and depth of cut.) 3. Surface roughness labels: Precisely measured Ra values obtained through white light interferometry.

To ensure rigorous evaluation of our augmentation framework while maintaining complete separation between training and testing phases, the entire dataset is randomly divided into two distinct subsets: The Training set consisted of 52 samples (81.25% of total data) used for HAS-CGAN and other CGANs training, and prediction model development. The testing set consisted of 12 samples (18.75% of the total data) that are completely excluded from any training process.

### 3.2 Data Generation and Results Analysis

After training the HAS-CGAN, the generator with a 3-layer fully connected network successfully produces synthetic force signals exhibiting remarkable similarity to real experimental data. Besides the HAS-CGAN, Auxiliary Conditional GAN can also use an auxiliary classifier in the discriminator to enforce label-conditioned generation, ensuring high alignment between generated samples and their corresponding labels. Also, the auxiliary loss mitigates mode collapse by encouraging diverse samples that match the target distribution. As for the WCGAN, it replaces Jenson-Shannon divergence with Earth Mover's Distance, effectively avoiding vanishing and mode collapse. Also, it enables more stable training, even with imbalanced data. **Fig. 2. a**. presents comparative time-domain waveforms of some of the generated force signals and the real force signals with the same surface roughness across five different CGANs, including our proposed HAS-CGAN, traditional Convolutional Conditional GAN, traditional Conditional GAN, Auxiliary Conditional GAN and Wasserstein Conditional GAN. The non-stationary nature of force signals in ultra-precision machining motivates our use of wavelet coherence (WC) analysis, which excels at evaluating time-localized similarities, particularly suited to periodic signals. What's more, WC can validate multi-scale fidelity, such as low and high-frequency components, making it suitable method to evaluate the similarity between generated signals and real signals. **Fig. 2. b**. is the wavelet coherence of those generated signals and real signals to quantitatively evaluate their similarity of. The results of wavelet coherence fall within 0 to 1. 0 means two signals are non-correlated and 1 means two signals who are totally co-rrelated and 1 means two signals are totally correlated. A higher value of it represents a higher similarity between generated signals and the real ones.

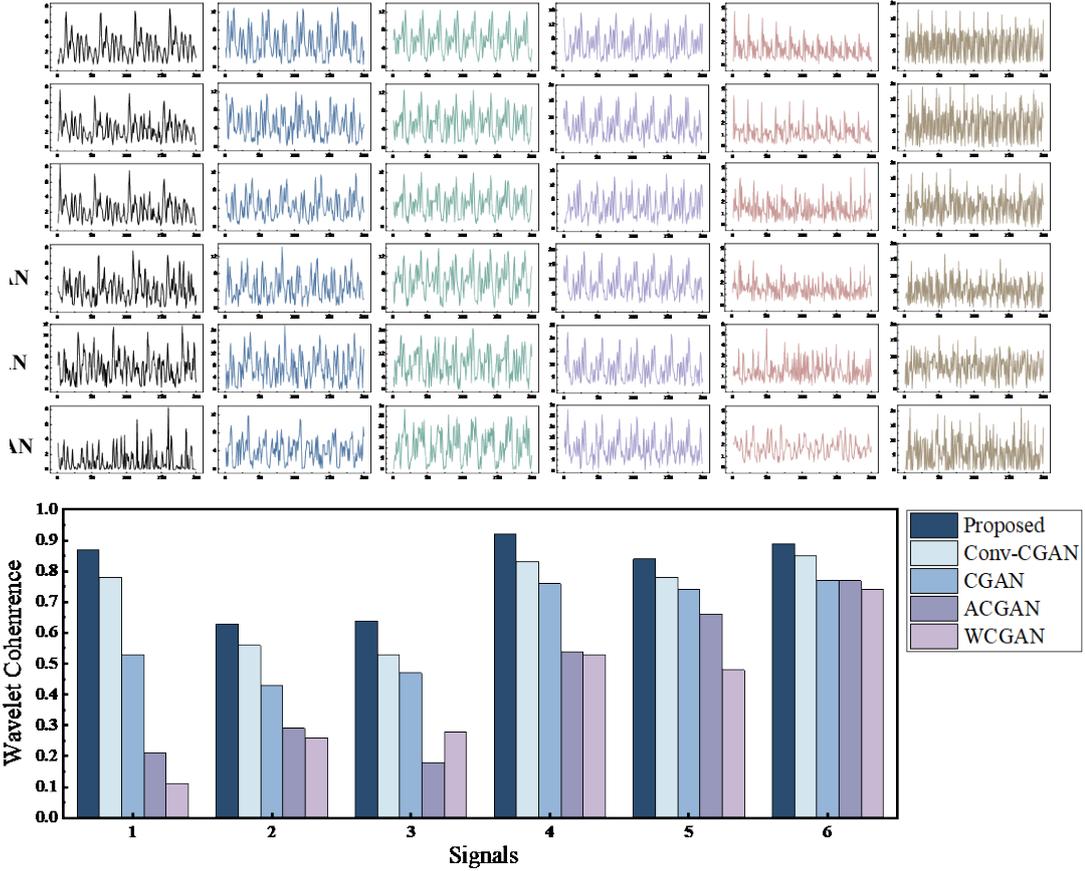

**Fig. 2.** The illustration of visual comparison and quantitative similarity computation of generated signals and real signals. **a.** The time domain waveform of force signals from generated samples and real samples under various conditional CGANs. **b.** The wavelet coherence of generated force signals and real force signals.

From **Fig. 2. a**, it can be seen observed in ACGAN and WCGAN generally underperform compared to the traditional CGAN, traditional Convolutional CGAN and our proposed method in general, as the range, shape and the detailed waveform of the generated signals show a lower fidelity with the original signals. This phenomenon is quantitatively confirmed in **Fig. 2. b** since the WC of ACGAN and WCGAN is no more than 0.8, and even smaller than 0.3 for signals 1, 2, and 3, which is quite lower than other CGANs. Most notably, our proposed method can generate signals most highly correlated with the original real signals with WC over 0.8. Both visual waveform comparison and quantitative WC analysis consistently indicate that time-domain and frequency-domain similarity between generated signals and the original signals from ACGAN and WCGAN is worse than other methods. This may be because the motivation of design ACGAN and WCGAN is excels at generating high-dimensional, complex data (e.g., images), but the simple structure of 1D signals (e.g., force waveforms) might not require its sophisticated Lipschitz constraints. Also, the complex architecture of ACGAN and WCGAN tends to overfit on small 1D dataset, whereas CGAN, with fewer parameters, demonstrates better generalization performance. What's more, the temporal dependencies in force signals (e.g., short-term autocorrelation in vibration signals) may be overlooked by the global optimization objectives of ACGAN or WCGAN, whereas the simpler architecture of CGAN can better capture local patterns.

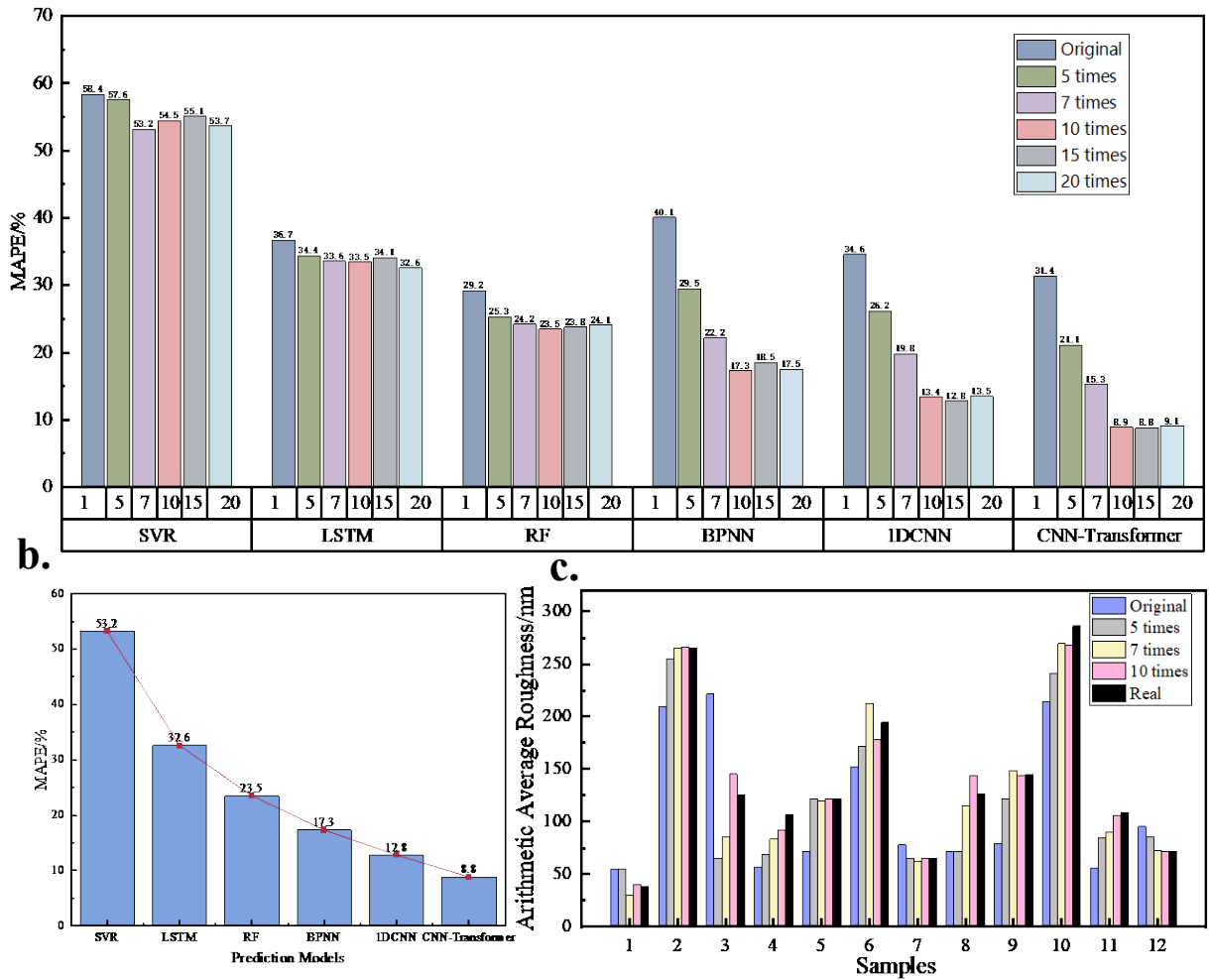

**Fig. 3.** The schematic illustration of the comparison of prediction results before and after data augmentation under various prediction methods and different times of augmentation datasets. **a.** Prediction performance under different times of training data from various prediction models. **b.** Best prediction results of various prediction methods. **c.** Prediction performance comparison from CNN-Transformer under different times of the augmented dataset.

Another phenomenon is further observed that Convolutional CGAN performs better than CGAN with dense layers. For example, it can be seen that in signal 4, the signal generated by Convolutional CGAN is more precise and closer to the original signal than that generated by CGAN with a dense layer. Or in signal 1, Convolutional CGAN demonstrates superior trend-fitting capability than CGAN with a dense layer. These observations align with WC results showing higher values for Convolutional CGAN. Lastly, as evidenced by the generated signals' visualization results, the traditional Convolutional CGAN demonstrates limited capability in high-frequency signal reconstruction. However, our proposed HAS-CGAN achieves superior results by penalizing high-frequency fitting error. For example, signals 5 and 6 are comparable high-frequency signals. No matter for traditional Convolutional CGAN, CGAN, ACGAN or WCGAN, it is hard for them to mimic the amplitude of the original signals, not to mention the waveform detail. However, when adding the spectral loss, which is computed in the frequency domain by transforming signals using Fourier Transform (FFT), aiming to keep the generation fidelity. The WC value for signal 5 and signal 6 can be seen in **Fig. 2. b,** that our proposed method with spectral loss punishment increases the coherence of generated signals and original signals to around 0.9.

**3.3 Results and Discussion of prediction accuracy after data augmentation**

To systematically investigate the influence of generated training sample on the final prediction accracy, comprehensive experiments are conducted with varying times of augmentation of original datasets, which is 5 times (210 samples), 7 times (364 samples),

10 times (520 samples), 15 times (780 samples), and 20 times samples (1040 samples) respectively. As illustrated in **Fig. 3. a**, several enlarged dataset are constructed by supplementing the original 56 real samples with generated samples at different scales. These values were carefully selected to represent incremental increases from 5 to 20 times the original dataset size, enabling a thorough examination of data augmentation effects.

Under SVR, LSTM and RF models, which use hand-made time-domain and frequency domain features as input, there is no big improvement before and after augmentation since the prediction accuracy, quantified using Mean Absolute Percentage Error (MAPE), is very large with around 50%, 35% and 25% respectively. This may be because the low correlation between hand-made time or frequency domain feature and surface roughness. And after feature extraction, the augmented samples only increase the redundant features instead of effective information for SVR, RF or LSTM.

However, under those end-to-end models which automatically extract features, such as BPNN, 1DCNN and CNN-Transformer models, experimental results demonstrate a clear trend in model prediction performance improvement with increasing datasets. MAPE shows significant enhancement from an initial 40.1% error down to 17.3% for BPNN, from 34.6% down to 12.8% for 1DCNN, and from 31.4% down to 8.8% for CNN-Transformer as the generated samples increase. This improvement follows a logarithmic pattern, with the most substantial gains occurring in the early stages of data augmentation. However, the performance curve exhibits a distinct plateau when the number of generated samples exceeds approximately 10 times (520 samples) the original real sample count. Beyond this critical threshold, the prediction accuracy of the best model (CNN-Transformer) stabilizes around 9% MAPE, suggesting that additional data generation provides diminishing returns. This phenomenon indicates the existence of an optimal augmentation range, beyond which further sample generation may not justify the associated computational costs.

More distinct comparisons and detailed visualization of these findings are presented in **Fig. 3. b** and **Fig. 3. c**. **Fig. 3. b** provides a direct comparison between the best model-CNN-Transformer and other prediction models, where the best result for each model is selected for comparison. It can be seen that CNN-Transformer with MAPE 8.8% outperforms other models, especially those models with hand-made features as input. **Fig. 3. c** provides a comparison of the prediction performance of CNN-Transformer across different sizes of training datasets. It clearly shows that prediction pink bars based on the 10 times augmented dataset (520 samples) exhibit remarkably similar height with the reference black bars denoting actual surface roughness measurements. This close correspondence demonstrates the model's enhanced predictive capability when trained with sufficient augmented data. Notably, even the 5 times augmentation case (gray bars, 260 samples) shows substantial improvement over the baseline model (blue bars) trained solely on the original 52 samples. The prediction results from **Fig. 3** show our proposed HAS-CGAN framework can effectively synthesize high-quality additional training samples, capture the complex relationships between machining parameters and surface roughness and improve prediction model performance despite severe data limitations, though there exists a practical limit to its benefits.

The experimental outcomes suggest that strategic sample generation can effectively overcome data scarcity limitations in neural network training, but requires careful consideration of the augmentation scale. The optimal balance between computational resource investment and model performance improvement appears to occur when the augmented dataset size is approximately 10 times the original sample count. This finding has important implications for practical applications where both prediction accuracy and resource efficiency are critical considerations.

## 4. Conclusion

This study investigated the application of conditional generative adversarial networks (CGANs) for data augmentation in ultra-precision machining (UPM) surface roughness prediction, addressing the critical challenge of limited training data. Through comparative analysis of various CGAN architectures—including traditional CGAN, convolutional CGAN, ACGAN, WCGAN, and our proposed HAS-CGA—we demonstrated that simpler architectures (CGAN/convolutional CGAN) outperform complex variants (ACGAN/WCGAN) for 1D force signal generation. This is attributed to: (1) the overfitting tendency of complex models on small datasets, (2) their inability to capture localized temporal dependencies in force signals, and (3) unnecessary Lipschitz constraints for low-dimensional data. Key findings reveal that our proposed HAS-CGAN achieves superior performance, with wavelet coherence (WC) values exceeding 0.9, by explicitly penalizing high-frequency fitting errors in the Fourier domain. The generated signals improved surface roughness prediction accuracy by 72% (MAPE reduction from 31.4% to 8.8% for CNN-Transformer) when augmenting the original 52-sample dataset with 10× synthetic data. However, diminishing returns were observed beyond this scale, highlighting an optimal augmentation threshold. These results establish CGAN-based data augmentation as a viable solution for UPM quality monitoring, with three broader implications: 1. Architecture Simplicity Matters: For 1D industrial signals, lightweight CGANs with spectral constraints outperform theoretically advanced variants. 2. Frequency-Aware Loss is Critical: Spectral loss preserves high-frequency features crucial for precision machining. 3. Real-Time Quality Control: The framework enables virtual metrology for UPM, reducing reliance on physical measurements.

Future work should explore hybrid models combining CGANs with physics-based simulations to further enhance data efficiency. This approach bridges the gap between limited experimental data and the demands of data-driven smart manufacturing.